\documentclass[11pt,a4paper]{article}
\usepackage[hyperref]{acl2019}
\usepackage{times}
\usepackage{latexsym}
\usepackage{float}
\usepackage{amsmath}
\usepackage[english]{babel}
\usepackage{graphicx}

\usepackage{url}
\usepackage{textcomp}

\title{Distributionally Robust Classifiers in Sentiment Analysis}

\author{Shilun Li \\
  Dept. of Mathematics\\
  Stanford University \\
  \texttt{shilun@stanford.edu} \\\And
  Renee Li \\
  Dept. of Computer Science\\
  Stanford University \\
  \texttt{reneeli@stanford.edu} \\\And
  Carina Zhang \\
  Dept. of Statistics\\
  Stanford University \\
  \texttt{carinaz@stanford.edu} \\}

\date{June, 2019}
\aclfinalcopy
\begin{document}
\maketitle

\section{Introduction}

Modern state-of-the-art sentiment classifiers have achieved high performance on experiments where training data and test data are from the same distribution. However, it has been shown that the classifiers do not generalize well when this assumption is broken. Recent research has developed a framework, Distributionally Robust Optimization (DRO), to offset the negative influence of shifting data distribution and make the underlying classification models perform better when there is a change in distribution \citep{duchi2018learning}. While heuristic methodologies have shown progress and were tested on synthetic datasets and computer vision datasets, we want to extend DRO to Natural Lanaguage Processing tasks, such sentiment analysis. 

In this project, we investigate how sentiment classification models integrated with DRO will behave under distributional shifts. To narrow down the scope, we will consider one form of distributional shift and understand how it affects our models. 

Our major contributions are:
\begin{itemize}
    \item successfully applying the DRO framework to develop machine learning models (specifically sentiment analysis within the field of NLP) when the test set's population and the training set's population do not overlap but are similar in nature;
    \item confirming through our experiments that our DRO model does improve performance on our test set with distributional shift from the training set;
    \item developing key observation that while there is no obvious pattern with small radii, the performance of our DRO model converges to the performance of baseline model as the radius of the Lp ball gets very large; thus, the best performing radius is usually somewhere in the middle.
\end{itemize}

\section{Related Works}

Recent research attempted to develop a model that can perform well across all unforeseen test distributions. While Duchi et. al. were able to develop and analyze a DRO framework that learns a model that provides good performance against perturbations to the data-generating distribution \citep{duchi2018learning}, we want to explore whether a model can be developed to be trained and tested on NLP datasets as well to show the increasing generality of the DRO framework.

Previous work in domain adaptation leads to development of models that receive data from one domain and are tested on a specified target. For example, there have been NLP-specific implementations of DRO \citep{oren2019distributionally, anonymous2020distributionally}.
Some researchers describe a framework that deals with a specific distributional shift, \textit{subpopulation shift}, in which the test distribution is a subpopulation of the training distribution. They have achieved better results than generic DRO and have shown that topics are an effective way to encode prior information about test distributions \citep{oren2019distributionally}. A stochastic optimizer is also introduced to a DRO framework that scales to large models and datasets, while improving robust accuracy at only a small cost in average accuracy \citep{anonymous2020distributionally}. 

Building upon these implementations, we will explore another type of distributional shift - when the training set's population and test set's population does not overlap but are similar in nature and have the same set of output labels. 

\section{Movie Review Datasets} 
We choose two movie review datasets, IMDB and Rotten Tomatoes, as our trainng set and test set. Since Rotten Tomatoes reviews and IMDB are similar data sets and generate either a "positive" or a "negatieve" sentiment output, Rotten Tomatoes can be considered as a shifted IMDB test set. 

\subsection{Training Set: IMDB Reviews}

We use Stanford AI Lab’s Large Movie Review Dataset \citep{Maas} as our training set. The dataset was collected from IMDB and contains roughly 50,000 movie reviews, in which 25,000 are positive and 25,000 are negative. The core dataset contains 50,000 reviews in total, we are using the 25,000 reviews from the training set. No more than 30 reviews are selected from the reviews pool of any single movie because reviews for the same movie tend to have correlated ratings. \\

\subsection{Test Set: Rotten Tomatoes Reviews}
We acquire labeled Rotten Tomatoes movie review dataset from the Stanford Treebank Project \citep{SocherEtAl2013}. Since we want to see how DRO performs on shifted a dataset, it seems most fitting for us to employ another large dataset of movie reviews as our test set. This dataset is split into train, dev, and test sets containing 8,544, 1,101, and 2,210 reviews respectively. We have used the pre-splitted test set as our test set. 

\section{Model and Algorithm}
This section describes our model and algorithm in detail. The baseline and oracle models and experiments are described in \emph{Section 5}.

\subsection{Advanced Metohd Model Description}
\begin{figure}[h]
    \centering
    \includegraphics[width=0.5\textwidth]{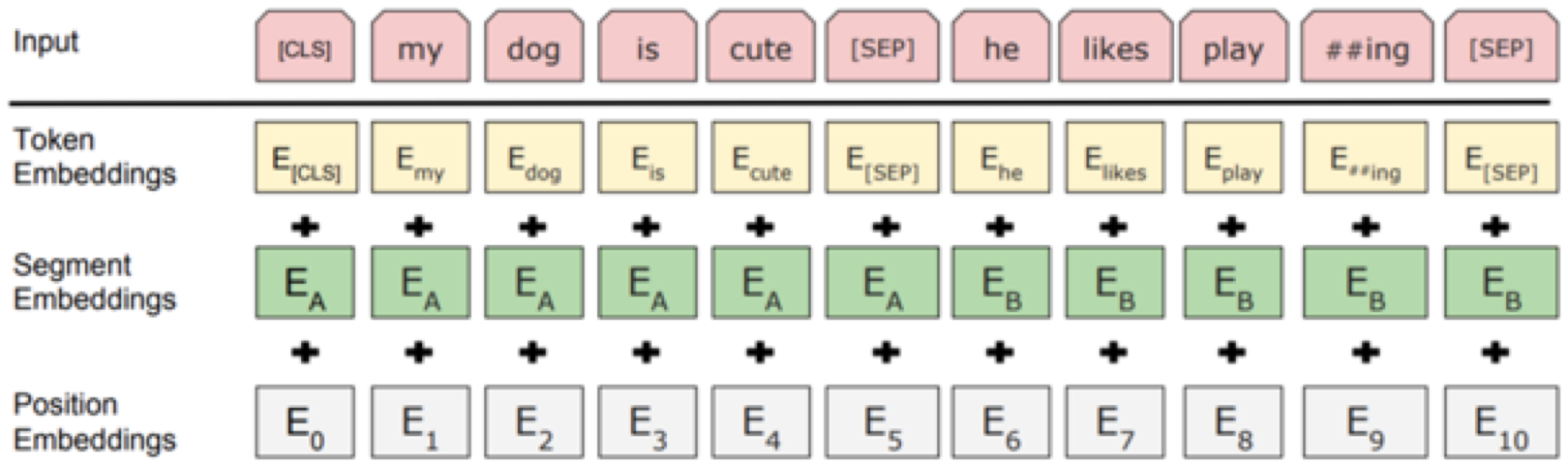}
    \includegraphics[width=0.5\textwidth]{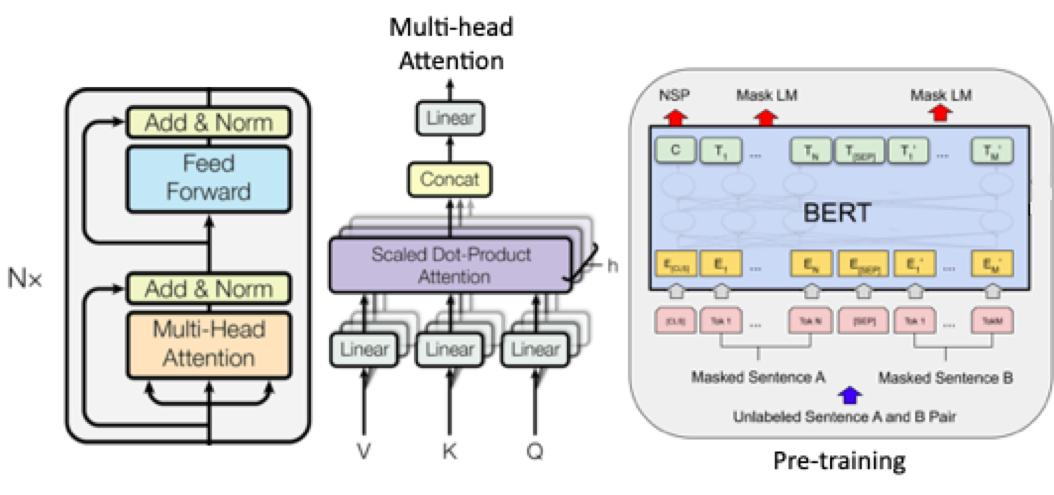}
    \caption{BERT’s key technical innovation is applying the bidirectional training of Transformer, a popular attention model, to language modelling, as shown in this example.}
    \label{fig:bert}
\end{figure}

We use PyTorch pre-trained BERT-base-cased model (referred to as BERT later) to produce word embeddings and use 2 layer bi-LSTM and linear classifier as our underlying classifier for sentiment classification. The pre-trained BERT is illustrated in Figure \ref{fig:bert} The  We then implement DROs on top of these classifiers and record the shifted test accuracy. Alternatives that we can potentially explore later would include SVM, Softmax, and Convolutional Neural Networks as classifiers.

\begin{figure}[h]
    \centering
    \includegraphics[width=0.2\textwidth]{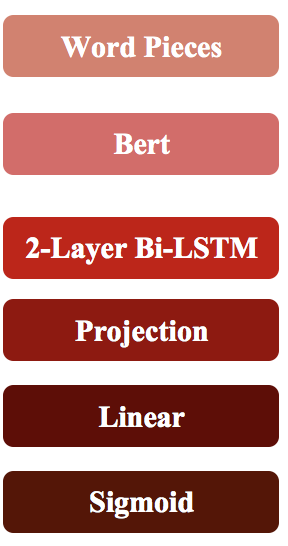}
    \caption{High level model architecture.}
    \label{fig:hlModel}
\end{figure}

In order to help our model better understand the reviews, we propose a model architecture (see Figure \ref{fig:hlModel}). As a comparison to just using DROs, we also did a controlled experiment where we trained BERT on IMDB dataset and tested on Rotten Tomatoes just using BERT. The architecture can be separated into 5 stages.

\begin{itemize}
    \item [a.] The first stage is the data pre-processing step. We have made some improvements to the pre-processing procedures from the one mentioned in our baseline model. We will discuss this step in more detail in Section \ref{experiment}. 
    \item [b.] In the second stage, we pass in word pieces from the dataset of movie reviews to the state-of-the-art language model for NLP -- BERT.
    \item [c.]
    At stage three, we pass the result we get from BERT into 2 bi-LSTM layers. While we are at the latter layer, we concatenated its forward and backward together.
    \item [d.]
    We project the result onto a $L_p$-ball or a simplex for the robust model.
    \item [e.]
    We apply dropout and then give the output to a linear classifier.
    \item [f.] 
    Finally, we use a sigmoid function to wrap up this model. Results and model comparison will be discussed in Section \ref{result}.
\end{itemize}

\subsection{Formulation of DRO}

Performance of machine learning models degrades significantly
on test sets that are different from what the model was trained on  because of its reliance on a priori fixed target distribution \citep{oren2019distributionally, duchi2018learning, anonymous2020distributionally}. To mitigate these challenges, we use a loss minimization framework that is explicitly robust to local changes in the data distribution. Concretely, let $\Theta \in R^d$ be the parameter space, $P_0$ be the data distribution, and $\ell(\theta; X)$ as the total loss function. When DRO models make predictions, rather than minimizing the average loss, $E[\ell(\theta; X)]$, we study the distributionally robust stochastic optimization problem, 
\begin{equation}
min_{\theta \in \Theta}  \{R_f(\theta; P_0):= sup_{Q <<P_0)} \{E_Q[\ell(\theta;X)]\}
\end{equation}

We call (1) the worst case risk, which captures how a model can performance in a worse-case situation. Optimizing worst-case situation is the crux of model robustness and generalizability. Since the worst-case risk upweights regions of X with high losses, it consequently optimizes performance on the tails. The model parameters that achieves the worst-case risk will be the parameters of the DRO model \cite{duchi2018learning}.

\subsection{Model workflow with examples}  
As mentioned in our proposal, we use two datasets, IMDB movie reviews dataset and Rotten Tomatoes movie reviews dataset, as training sets and test sets.

The pre-processing step is explained in details in \emph{Section 5} in our demonstration of our experiment.

We tokenize the sentences to feed into BERT. Since BERT has a constraint of a maximum token length of 512, we only take the first 512 tokens of a sentence to represent that sentence. After data processing, each movie review can be represented as a list of BERT token indexes of length $d$, where $d=512$ in our context.

Since neural networks spill out an unbounded intermediate output, $y'$, we need to apply sigmoid function $\hat{y} = \frac{1}{1 + e^{-y'}}$ to bound have a probability normalized between $(0, 1).$

Since sentiment classification is a binary classification problem, we investigate distributional shifts via a binary classification experiment using the binary cross entropy with sigmoid, defined as $\ell (\theta; x, y) = \frac{1}{N} \Sigma_{i = 1}^N y_i\cdot log(\hat{y_i}) + (1 - y_i) \cdot log({1 - \hat{y_i}}).$

As a \textbf{concrete example}, our training data from IMDB will look like this: "If you like original gut wrenching laughter you will like this movie. If you are young or old then you will love this movie, hell even my mom liked it. Great Camp!!!" and is labeled as "positive". Our test data (after pre-processing) from Rotten Tomatoes for the model would look like this: "Take Care of My Cat offers a refreshingly different slice of Asian cinema ." The correct prediction should be be "1" because it is a sentence with a "positive" sentiment.

Last but not least, since PyTorch's LSTM requires that the batch is sorted by decreasing token length, we complete the sorting of each batch while collecting the batches so that when each batch is passed into the biLSTM layer, it has already been sorted in the order we need.

The workflow with example inputs/outputs and processes is visualized in \ref{fig:model} below.

\begin{figure}[h]
    \centering
    \includegraphics[width=0.5\textwidth]{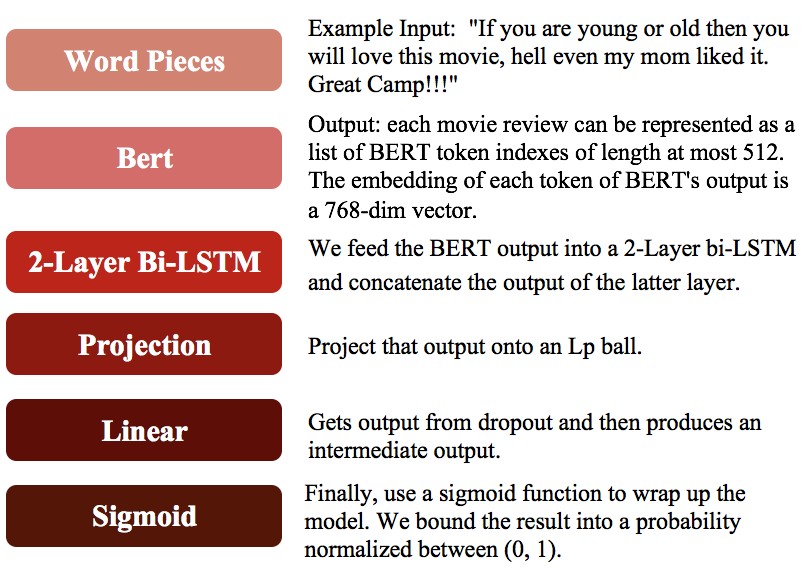}
    \caption{Model Architecture with example inputs/outputs and process.}
    \label{fig:model}
\end{figure}

\subsection{Potential challenges}
\begin{enumerate}
    \item Since we are dealing with truly unforeseen test data in our experiment, how much our model can improve from the baseline is not guaranteed. 
    \item It is hard to mathematically define the distributional shift explored in our experiment. It is obvious that IMDB reviews and Rotten Tomatoes reviews are similar datasets for a human being; however, it is hard to quantify the magnitude of the shift, which makes the results of our project hard to generalize beyond movie reviews. 
\end{enumerate}

\section{Experiment}\label{experiment}
In this section, we first introduce the evaluation metrics we use for the models. Then we introduce the methods we adopt to preprocess the text information. Lastly, we describe the experiments we conducted in our project.


\subsection{Evaluation Design and Metrics}
\begin{enumerate}
    \item Independent Variables: Test Accuracy
    \item Dependent Variables: 
    \begin{enumerate}
        \item Base classifier: the model used to accomplish sentiment classification task.
        \item Test set distributional shift: the distribution of data that we expose the models to in testing.
    \end{enumerate}
\end{enumerate}

\subsection{Data pre-processing}
For the training set, we use the labeled IMBD train set for this project. In the labeled training set, a negative review has a score $\leq$ 4 out of 10, and a positive review has a score $\geq$ 7 out of 10. Reviews with neutral ratings (with scores of 5 or 6)  are not included in the training sets.

For the test set, since our task is binary classification, review scores have a range from 0 to 4. A negative review has a score $<$ 2, and a positive review has a score $>$ 2. Reviews with score equal to 2 are not included in out dataset.

We filter the dataset as described above and output the sentences with sentiment labels, where $1$ indicates positive sentiment and $0$ indicates nagetive sentiment. We output the parsed dataset to a .csv file, which speeds up the process of data loading during training.

As a \textbf{concrete example}, our input data (after pre-processing) for the model would look like this: "Take Care of My Cat offers a refreshingly different slice of Asian cinema ." And the output would be "1" because it is a sentence with a "positive" sentiment.

\subsection{Baseline Experiment} \label{naiveapproach}
We implement GloVe model, a relatively naive sentiment classifier, as our baseline model \citep{pennington2014glove}. We trained our GloVe on IMBD data set and tested on Rotten Tomatos test set. We use test accuracy obtained by GloVe without applying DRO as our baseline accuracy, which is 77.7\%. 

\subsection{Oracle Experiment}
We choose the best performing model in terms of in-distribution test accuracy on the labeled Rotten Tomatoes reviews dataset, ALBERT, which is a lighter version of BERT model with fewer parameters with regularization. This model achieved an in-distribution test accuracy of 97.1\% \citep{lan2019albert}.

\subsection{Advanced Method Experiment}
\begin{enumerate}
    \item Encoding: We use contextualized word embedding produced by BERT from word pieces to encode the text inputs, which are movie reviews stored as a string. Since fine-tuning the model requires too much time, we freeze the pre-trained BERT to get a fixed representation of words to get the vector presentation to feed into DRO. 
    \item Training: We train using a subset of the different types of classifiers mentioned above on the IMDB data set after the pre-process procedures as mentioned above.
    \item DRO Implementaion: We have attempted the implementation of several DROs to our classifiers. In particular, we have tried projecting on an $L_p-$ball where $p=1,2 \text{ or }4$ and a simplex.
    \item Testing: We test our DRO-integrated models (i.e. robust models) on the Rotten Tomatoes test set and compare the results between in-distribution test accuracy and shifted test accuracy and evaluate using the metrics explained below.
\end{enumerate}

\section{Results and Discussion}
We implemented two non-robust classification models for comparison purposes and two robust models with $L_1$, $L_2$, $L_4$ and Simplex with give different radii, $R = 1, 2, 5, 10, 15$. We report test accuracy for each model.

\subsection{Non-Robust Model Performance}
We implemented our classification model by the procedure we described in the previous sections without implementing DRO. Thus, this classification model based on pre-trained BERT and linear classifier can be viewed as a baseline of how classification models perform under distributional shifts. The results of the model is shown in Table-\ref{nonrobust}. We can see from the table that there exists a significant drop in performance (around $10\%$) when the test set comes from a different population.
    \begin{center}
    \begin{table}
    \caption{Results of non-robust model performance}
    \begin{tabular}{||c c c||} 
    \hline
    Train set & Test Set & Test Accuracy  \\ 
    \hline\hline
    IMDB & IMDB & 83.60\% \\ 
    \hline
    IMDB & R. Tomatoes & 73.64\%\\ 
    \hline\hline
    \end{tabular}
    \label{nonrobust}
    \end{table}
    \end{center}

\subsection{DRO Models Performance}
\begin{figure}[h]
    \centering
    \includegraphics[width=0.5\textwidth]{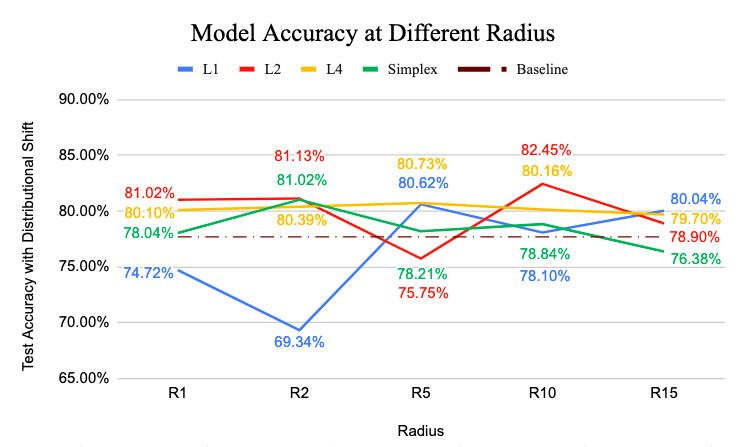}
    \caption{Models' out-of-distribution accuracy with difference configurations}
    \label{result}
\end{figure}

We added the DRO implementation described in previous sections with different model configurations, namely  Simplex, $L_1$, $L_2$, $L_4$ projections with radii $R = 1, 2, 5, 10, 15$. We report test accuracy without distributional shift (referred as in-distribution test accuracy, namely trained on IMDB and tested on IMDB) and test accuracy with distributional shift (referred as out-of-distribution test accuracy, namely trained on IMDB and tested on Rotten Tomatoes) for each model. The overall performance is reflected in Table-\ref{l1} to Table-\ref{simplex} and plotted in Figure-\ref{result}. 

\begin{table}
\caption{DRO Model with $L_1$ Projection}
\begin{tabular}{||c c c||} 
    \hline
    Radius & \begin{tabular}{@{}c@{}}In-distribution \\Test Accuracy \end{tabular}  & \begin{tabular}{@{}c@{}}Out-of-distribution \\Test Accuracy \end{tabular} \\ 
    \hline\hline
    1 & 80.65\% & 74.72\% \\
    \hline
    2 & 75.10\% & 69.34\% \\
    \hline
    5  & 84.55\% & 80.62\% \\
    \hline
    10 & 84.25\% & 78.10\% \\
    \hline
    15 & 85.00\% & 80.04\% \\
    \hline
\end{tabular}
\label{l1}
\end{table}

\begin{table}
\caption{DRO Model with $L_2$ Projection}
\begin{tabular}{||c c c||} 
    \hline
    Radius & \begin{tabular}{@{}c@{}}In-distribution \\Test Accuracy \end{tabular}  & \begin{tabular}{@{}c@{}}Out-of-distribution \\Test Accuracy \end{tabular} \\ 
    \hline\hline
    1 & 84.90\% & 81.02\% \\
    \hline
    2 & 84.85\%	& 81.13\% \\
    \hline
    5  & 83.45\% &75.75\% \\
    \hline
    10 & 85.60\%	& 82.45\% \\
    \hline
    15 & 84.90\% &	78.90\%\\
    \hline
\end{tabular}
\label{l2}
\end{table}

\begin{table}
\caption{DRO Model with $L_4$ Projection}
\begin{tabular}{||c c c||} 
    \hline
    Radius & \begin{tabular}{@{}c@{}}In-distribution \\Test Accuracy \end{tabular}  & \begin{tabular}{@{}c@{}}Out-of-distribution \\Test Accuracy \end{tabular} \\ 
    \hline\hline
    1 & 84.90\%	  & 80.10\% \\
    \hline
    2 & 85.50\%  & 	80.39\% \\
    \hline
    5  & 85.00\% & 	80.73\% \\
    \hline
    10 & 84.95\% & 	80.16\%\\
    \hline
    15 & 85.10\% & 	79.70\%\\
    \hline
\end{tabular}
\label{l4}
\end{table}

\begin{table}
\caption{DRO Model with Simplex Projection}
\begin{tabular}{||c c c||} 
    \hline
    Radius & \begin{tabular}{@{}c@{}}In-distribution \\Test Accuracy \end{tabular}  & \begin{tabular}{@{}c@{}}Out-of-distribution \\Test Accuracy \end{tabular} \\ 
    \hline\hline
    1 & 84.45\% & 	78.04\% \\
    \hline
    2 & 83.60\% & 	81.02\%\\
    \hline
    5  & 84.70\% & 	78.21\% \\
    \hline
    10 & 84.35\%	 & 78.84\% \\
    \hline
    15 & 84.20\%	 & 76.38\%\\
    \hline
\end{tabular}
\label{simplex}
\end{table}

\subsection{Analysis and Interpretation of Experimental Results}

Comparing results from our DRO models to that of the non-rost model, our observations align with the assumption that DRO performs better under distributional shifts. When exposed to shifted data, the certified test accuracy of our model decreases from above 90\% to 77\%. Thus, this result has shown than when models are exposed to distributional shift (in our context, the population changes from IMDB and Rotten Tomatoes). We can infer that the current state-or-art sentiment classifier is not robust under distributional shifts. 

One potential cause of the drop in performance is that empirically, there exists an inherent trade-off between robustness and accuracy. Increasing robustness requires more relaxation and thus may result in lower certified test accuracy for a particular set of perturbations. Since the three models studied in our paper are tuned to a specific fixed, known family of distortions, the generalizability of these models decreases. 

An observation that aligns with this potential cause of drop in performance is the observation that test accuracy and radius does not have a linear relationship. Intuitively, the smaller R is, the more "relaxed" the optimization problem is. Larger R might lead to convergence to in-distribution results. However, since DRO models are not tuned to fit specifically to this specific distributional shift, there is a robustness-accuracy trade off. Thus, a radius value in the middle usually produce the best results. 

We identified two potential sources of error. First, we see that sometimes our model's performance dropped below the baseline. Because of the limitation in time and computational power, we did not have time to run every experiment several times and record the average result. Thus, randomness causes the variance in model performance. Second, the challenge lies in the fact that the effect of projection on the original input in a natural language setting is more ambiguous than images. Since it is hard to mathematically define our distributional shift, it is also hard to precisely define the projection's effect on sentences. Some of the projection methods, such as $L_1$, is not suitable in this setting. However, we included this model configuration for the sake of completeness. 

\section{Conclusion}
In this work, we propose sentiment classification models integrated with DRO to improve model performance on datasets with distributional shifts. To narrow down the scope, we consider one form of distributional shift (from IMDb dataset to Rotten Tomatoes dataset). We evaluate our model's performance based on test accuracy. We conduct experiments and show that the model's performance does not have a linear relationship with the radius of the Lp balls we project onto; instead, the performance gradually converges to baseline performance with the increase of radius. We found that out of Simplex, $L_1$, $L_2$, and $L_4$ projections, $L_4$ had the best overall performance. 

The major contributions of this work include applying the DRO framework to machine learning models (specifically sentiment analysis within the field of NLP) when the test set's population and the training set's population do not overlap but are similar in nature. Meanwhile, we have confirmed through our experiments that our DRO model does improve performance on our test set with distributional shift from the training set.

\subsection{Future Work}
While we successfully accomplished our task detailed in the Introduction section through this project, there are potential directions we want to explore and questions we want to answer in the future. Overall, we want to look into whether we could add more diversity to our distributional shifts beyond a total population shift. More specifically, we can experiment with mixing up IMDB and Rotten Tomatoes datasets with different ratios to have different distributions to test on. We are also interested in further exploring with datasets for which we can better quantify the underlying distributional shift.

\section*{Acknowledgement} 
We would like to acknowledge the help of Professor Percy Liang, Professor Dorsa Sadign, and Professor John Duchi for brainstorming the project idea, Dr. Hongseok Namkoong for the initial codebase. We would also like to thank our amazing teaching assistant Haoshen Hong for providing helpful suggestions and feedback as well as effective solutions to tackle the problem.


{\small
\bibliographystyle{acl_natbib}
\bibliography{acl2019}
}

\section*{Supplementary Material}
Link to github: \url{https://github.com/Shilun-Allan-Li/CS221-Fall2019}
\end{document}